\documentclass[preprint,12pt]{elsarticle}

\usepackage{amssymb}
\usepackage{algorithm,algorithmic}
\usepackage{color}
\usepackage{setspace}

\journal{}

\begin{document}

\begin{frontmatter}



\title{A comprehensive review of firefly algorithms}


\author[mb]{Iztok ̃Fister\corref{cor1}}		
\ead{iztok.fister@uni-mb.si}
\author[mb]{Iztok ̃Fister Jr.}				
\ead{iztok.fister2@uni-mb.si}
\author[uk]{Xin-She ̃Yang}					
\ead{x.yang@mdx.ac.uk}
\author[mb]{Janez ̃Brest}					
\ead{janez.brest@uni-mb.si}

\cortext[cor1]{Corresponding author}

\address[mb]{University of Maribor, Faculty of electrical engineering and computer science, Smetanova 17, 2000 Maribor, Slovenia.}
\address[uk]{School of Science and Technology, Middlesex University, London NW4 4BT, UK.}

\begin{abstract}

The firefly algorithm has become an increasingly important tool of Swarm Intelligence that has been applied in almost all areas of optimization, as well as engineering practice. Many problems from various areas have been successfully solved using the firefly algorithm and its variants. In order to use the algorithm to solve diverse problems, the original firefly algorithm needs to be modified or hybridized. This paper carries out a comprehensive review of this living and evolving discipline of Swarm Intelligence, in order to show that the firefly algorithm could be applied to every problem arising in practice. On the other hand, it encourages new researchers and algorithm developers to use this simple and yet very efficient algorithm for problem solving. It often guarantees that the obtained results will meet the expectations. \\
 
Citations details: I. Fister, I. Fister Jr., X.-S. Yang, and J. Brest, “A comprehensive review of firefly algorithms,” Swarm and Evolutionary Computation, vol. 13, pp. 34-46, 2013. \\

\end{abstract}


\end{frontmatter}


\section{Introduction}
\label{Introduction}

Swarm Intelligence (SI) belongs to an artificial intelligence discipline (AI) that became increasingly popular over the last decade~\cite{Blum:2008}. It is inspired from the collective behavior of social swarms of ants, termites, bees, and worms, flock of birds, and schools of fish. Although these swarms consist of relatively unsophisticated individuals, they exhibit coordinated behavior that directs the swarms to their desired goals. This usually results in
the self-organizing behavior of the whole system, and collective intelligence or swarm intelligence is in essence the
self-organization of such multi-agent systems, based on simple interaction rules. This coordinated behavior is performed due to interaction between individuals, for example, termites and worms are able to build sophisticated nests, whilst ants and bees also use this collective behavior when searching for food. Typically, ants interact with each other via chemical pheromone trails in order to find the shortest path between their nest and the food sources. In a bee colony, the role of informer is played by so-called scouts, i.e., individual bees that are responsible for searching for new promising areas of food sources. Here, the communication among bees is realized by a so-called 'waggle dance', through which the bee colony is directed by scouts. During this discovery of the new food sources, a trade-off between exploration (the collection of new information) and exploitation (the use of existing information) must be performed by the bee colony~\cite{Beekman:2008}. That is, the bee colony must be aware when to exploit existing food sources and when to look for new food sources so as to maximize the overall nectar intake while minimizing the overall foraging efforts.

The swarm of individuals shows collective behavior; for example, where to forage, when to reproduce, where to live, and how to divide the necessary tasks amongst the available work force~\cite{Beekman:2008}. In fact, these decisions are made in a decentralized manner by individuals based on local information obtained from interactions with their intermediate environments.

Swarm intelligence refers to a research field that is concerned with a collective behavior within self-organized and decentralized systems. This term was probably first used by Beni~\cite{Beni:1989} in the sense of cellular robotic systems consisting of simple agents that organize themselves through neighborhood interactions. Recently, methods of swarm intelligence are used in optimization, the control of robots, and routing and load balancing in new-generation mobile telecommunication networks, demanding robustness and flexibility. Examples of notable swarm-intelligence optimization methods are ant colony optimization (ACO)~\cite{Dorigo:1999} \cite{Korosec:2012}, particle swarm optimization (PSO)~\cite{Kennedy:1999}, and artificial bee colony (ABC)~\cite{Karaboga:2007} \cite{Fister:2012}. Today, some of the more promising swarm-intelligence optimization techniques include the firefly algorithm (FA)~\cite{yang2008afirefly} \cite{gandomi2013metaheuristic} \cite{yang2013metaheuristic} \cite{sahab2013metaheuristic}, cuckoo-search~\cite{Yang:2009}, and the bat algorithm~\cite{Yang:2010}, while new algorithms such as the krill herd bio-inspired optimization algorithm~\cite{Gandomikrillherd} and algorithms for clustering~\cite{hatamlou2012combined} \cite{hatamlou2012black} also emerged recently.

FA is one of the recent swarm intelligence methods developed by Yang~\cite{yang2008afirefly} in 2008 and is a kind of stochastic, nature-inspired, meta-heuristic algorithm that can be applied for solving the hardest optimization problems (also NP-hard problems). This algorithm belongs to stochastic algorithms. This means, it uses a kind of randomization by searching for a set of solutions. It is inspired by the flashing lights of fireflies in nature. Heuristic means `to find' or `to discover solutions by trial and error'~\cite{yang2008afirefly}. In fact, there is no guarantee that the optimal solution will be found in a reasonable amount of time. Finally, meta-heuristic means 'higher level', where the search process used in algorithms is influenced by certain trade-off between randomization and local search~\cite{yang2008afirefly}. In the firefly algorithm, the `lower level' (heuristic) concentrates on the generation of new solutions within a search space and thus, selects the best solution for survival. On the other hand, randomization enables the search process to avoid the solution being trapped into local optima. The local search improves a candidate solution until improvements are detected, i.e., places the solution in local optimum.

Each meta-heuristic search process depends on balancing between two major components: exploration and exploitation~\cite{crepinsek2011analysis}. Both terms were defined implicitly and are affected by the algorithm's control parameters. In the sense of the natural bee colony, the actions of exploration and exploitation has yet to be explained. For the meta-heuristic algorithms~\cite{Tashkova:ecological}, the exploration denotes the process of discovering the diverse solutions within the search space, whilst exploitation means focusing the search process within the vicinities of the best solutions, thus, exploiting the information discovered so far.

Note that FA is population-based. The population-based algorithms have the following advantages when compared with single-point search algorithms~\cite{Prügel:2010}:
\begin{itemize}
\item Building blocks are put together from different solutions through crossover.
\item Focusing a search again relies on the crossover, and means that if both parents share the same value of a variable, then the offspring will also have the same value of this variable.
\item Low-pass filtering ignores distractions within the landscape.
\item Hedging against bad luck in the initial positions or decisions it makes.
\item Parameter tuning is the algorithm's opportunity to learn good parameter values in order to balance exploration against exploitation.
\end{itemize}

The rest of this section will discuss briefly the characteristics of fireflies that have served as an inspiration for developing the firefly algorithm.

The main characteristic of fireflies is their flashing light. These lights have two fundamental functions: to attract mating partners and to warn potential predators. However, the flashing lights obey more physical rules. On the one hand, the light intensity \textit{I} decreases as the distance \textit{r} increases according to the term $I \propto 1/r^{2}$. This phenomenon inspired Yang~\cite{yang2008afirefly} to develop the firefly algorithm. On the other hand, the firefly acts as an oscillator that charges and discharges (fires) the light at regular intervals, i.e., at $\theta = 2\pi$. When the firefly is placed within the vicinity of another firefly, a mutual coupling occurs between both fireflies. This behavior of fireflies especially inspired the solution of graph coloring problems. On this basis, a distributed graph coloring algorithm was developed by Lee~\cite{lee:2010}. Recently, the similar and greater researched behavior of Japanese tree frogs inspired Hern\'{a}ndez and Blum~\cite{Hernandez:2012} into developing a more useful distributed graph coloring algorithm. Therefore, the further development of the algorithm based on the oscillatory behavior of fireflies has diminished. Therefore, in this paper we are focused on Yang's firefly algorithm.

An aim of this paper is twofold: to present areas where FA has been successfully applied, and thus to broaden the range of its potential users. The structure of this paper is as follows: Section 2 discusses the biological foundations of the firefly algorithm. Main characteristics of this algorithm are then exposed, and finally, the algorithmic structure is presented. Section 3 provides an extensive review of application areas to which this algorithm has already been applied. Let us mention only the most important areas of its application: continuous, combinatorial, constraint and multi-objective optimization, and optimization in  dynamic and noisy environments. Beside optimization, it is applicable for solving classification problems arose in areas like machine learning, data mining, and neural network. Additionally, many applications cover an area of engineering applications and solve real-world problems. Section 4 brings the discussion of FA behavior and directions for further development of this algorithms are covered. This paper concludes with an overview of the work that has been performed within the discipline of swarm intelligence.

\section{Firefly algorithm}

\subsection{Biological foundations}

Fireflies ({\tt Coleoptera: Lampyridae}) are amongst the most charismatic of all insects, and their spectacular courtship displays have inspired poets and scientists alike~\cite{lewis2008flash}. Nowadays, more that 2,000 species exist worldwide. Usually, fireflies live in a variety of warm environments and they are most active in summer nights. A lot of researchers have studied firefly phenomena in nature and there exist numerous papers
researching fireflies, for example, \cite{de1987firefly, brasier1989optimized, strehler1952firefly, deluca2006firefly, seliger1960spectral}.

Fireflies are characterized by their flashing light produced by biochemical process bioluminescence. Such flashing light may serve as the primary courtship signals for mating. Besides attracting mating partners, the flashing light may also serve to warn off potential predators. Note that in some firefly species some adults are incapable of bioluminescence. These species attract their mates due to pheromone, similarly to ants.

In fireflies, bioluminescent reactions take place from light-producing organs called lanterns. The most bioluminescent organisms provide only slowly modulated flashes (also glows). In contrast, adults in many firefly species are able to control their bioluminescence in order to emit high and discrete flashes. The lanterns' light-production is initialized by signals originating within the central nervous system of firefly.

Most firefly species rely on bioluminescent courtship signals. Typically, the first signalers are flying males, who try to attract flightless females on the ground. In response to these signals, the females emit continuous or flashing lights. Both mating partners produce distinct flash signal patterns that are precisely timed in order to encode information like species identity and sex. Females are attracted according to behavioral differences in the courtship signal. Typically, females prefer brighter male flashes. It is well-known that the flash intensity varies with the distance from the source. Fortunately, in some firefly species females cannot discriminate between more distant flashes produced by stronger light source and closer flashes produced by weaker light sources.

Firefly flash signals are highly conspicuous and may therefore deter a wide variety of potential predators. In the sense of natural selection~\cite{darwin:1859}, where only the strongest individual can survive, flash signals evolve as defense mechanisms that serve to warn potential predators.

Two features are characteristics for swarm intelligence: self-organization and decentralized decision making. Here, autonomous individuals live together in a common place as, for example, bees in hives, ants in anthills, etc. In order to live in harmony, some interaction or communication is needed amongst group members who live together (sociality). In fact, individuals within a group cannot behave as if they are solitary, but must adapt to the overall goals within the groups. The social life of fireflies is not just dedicated to foraging, but more to reproduction. These collective decisions are closely connected with the flashing light behavior that served as the main biological foundation for developing the firefly algorithm.

\subsection{Structure of the firefly algorithm}

As mentioned in Section~\ref{Introduction}, this paper focuses on Yang's~\cite{yang2008afirefly} implementation of the firefly algorithm. This algorithm is based on a physical formula of light intensity $I$ that decreases with the increase of the square of the distance $r^{2}$. However, as the distance from the light source increases, the light absorption causes that light becomes weaker and weaker. These phenomena can be associated with the objective function to be optimized. As a result, the base FA can be formulated as illustrated in Algorithm~\ref{alg:prog1}.

\begin{algorithm}[htb]
\caption{Pseudo code of the base Firefly algorithm}
\label{alg:prog1}
\footnotesize
\begin{algorithmic}[1]
\STATE $t = 0; s^{*} = \emptyset; \gamma = 1.0;$ \ \ \ \ \ \ \ \ // initialize: gen.counter, best solution, attractiveness
\STATE $P^{(0)}$ = InitializeFA(); \ \ \ \ \ \ \ \ // initialize a population
\WHILE {($t <$MAX\_FES)} 							
\STATE \ \ \ \ $\alpha^{(t)}$= AlphaNew(); \ \ \ \ \ \ // determine a new value of $\alpha$
\STATE \ \ \ \ EvaluateFA($P^{(t)}, f(s)$); \ \ \ \ \ \ // evaluate $s$ according to $f(s)$
\STATE \ \ \ \ OrderFA($P^{(t)}, f(s)$);	 \ \ \ \ \ \ // sort $s$ according to $f(s)$
\STATE \ \ \ \ $s^{*}$ = FindTheBestFA($P^{(t)}, f(s)$);\ \ // determine the best solution
\STATE \ \ \ \ $P^{(t+1)}$ = MoveFA($P^{(t)}$);\ \ \ \ \ \ \ \ // vary the attractiveness accordingly
\STATE \ \ \ \ $t = t+1$;
\ENDWHILE
\end{algorithmic}
\end{algorithm}

Some flashing characteristics of the fireflies are idealized in order to formulate the FA, as follows:

\begin{itemize}
\item All fireflies are unisex.
\item Their attractiveness is proportional to their light intensity.
\item The light intensity of a firefly is affected or determined by the landscape of the fitness function.
\end{itemize}

The population of fireflies is initialized by the `InitializeFA' function. Typically, this initialization is performed randomly. The firefly search process comprises the inside of the \textbf{while} loop (lines 3-10 in Algorithm~\ref{alg:prog1}) and is composed of the following steps: Firstly, the `AlphaNew' function is dedicated to modify the initial value of parameter $\alpha$. Note that this step is optional in the firefly algorithm. Secondly, the `EvaluateFA' function evaluates the quality of the solution. The implementation of a fitness function $f(s)$ is performed inside this. Thirdly, the `OrderFA' function sorts the population of fireflies according to their fitness values. Fourthly, the `FindTheBestFA' function selects the best individual in population. Finally, the `MoveFA' function performs a move of the firefly positions in the search space. Note that the fireflies are moved towards the more attractive individuals.

The firefly search process is controlled by the maximum number of fitness function evaluations MAX\_FES.

\subsection{Characteristics of the firefly algorithm}

In order to design FA properly, two important issues need to be defined: the variation of light intensity and the formulation of attractiveness. These two issues enable developers to tailor different firefly algorithms in such a manner that they are best suited to the demands of the problems to be solved. In the standard firefly algorithm, the light intensity $I$ of a firefly representing the solution $s$ is proportional to the value of fitness function $I(s) \propto f(s)$, whilst the light intensity $I(r)$ varies according to the following equation:
\begin{equation}
\label{eq:intensity}
I(r)=I_{0}e^{-\gamma r^{2}},
\end{equation}
where $I_{0}$ denotes the light intensity of the source, and the light absorption is approximated using the fixed light absorption coefficient $\gamma$. The singularity at $r=0$ in the expression $I/r^{2}$ is avoided by combining the effects of the inverse square law and an approximation of absorption in Gaussian form. The attractiveness $\beta $ of fireflies is proportional to their light intensities $I(r)$. Therefore, a similar equation to Eq.~(\ref{eq:intensity}) can be defined, in order to describe the attractiveness $\beta$:
\begin{equation}
\label{eq:attractiveness}
\beta=\beta_{0}e^{-\gamma r^{2}},
\end{equation}
where $\beta_{0}$ is the attractiveness at $r=0$. The light intensity $I$ and attractiveness $\beta$ are in some way synonymous. Whilst the intensity  is referred to as an absolute measure of emitted light by the firefly, the attractiveness  is a relative measure of the light that should be seen in the eyes of the beholders and judged by other fireflies (Yang, 2008).

The distance between any two fireflies $s_{i}$ and $s_{j}$ is expressed as Euclidean distance by the base firefly algorithm, as follows:
\begin{equation}
\label{eq:distance}
r_{ij}= \parallel s_{i}-s_{j} \parallel = \sqrt{\sum_{k=1}^{k=n}(s_{ik}-s_{jk})^{2}},
\end{equation}
where $n$ denotes the dimensionality of the problem. The movement of the $i$-th firefly is attracted to another more attractive firefly $j$. In this manner, the following equation is applied:
\begin{equation}
\label{eq:move}
s_{i} = s_{i}+\beta_{0}e^{-\gamma r_{ij}^{2}}(s_{j}-s_{i})+\alpha \epsilon_{i},
\end{equation}
where $\epsilon_{i}$ is a random number drawn from Gaussian distribution. The movements of fireflies consist of three terms: the current position of $i$-th firefly, attraction to another more attractive firefly, and a random walk that consists of a randomization parameter $\alpha$ and the random generated number from interval $[0,1]$. When $\beta_{0}=0$ the movement depends on the random walk only. On the other hand, the parameter $\gamma$ has a crucial impact on the convergence speed. Although the value of this parameter can theoretically capture any value from interval $\gamma \in [0, \infty)$, its setting depends on the problem to be optimized. Typically, it varies from 0.1 to 10.

In summary, FA is controlled by three parameters: the randomization parameter $\alpha$, the attractiveness $\beta$, and the absorption coefficient $\gamma$. According to the parameter setting, FA distinguishes two asymptotic behaviors. The former appears when $\gamma \rightarrow 0$ and the latter when $\gamma \rightarrow \infty$. If $\gamma \rightarrow 0$, the attractiveness becomes $\beta = \beta_{0}$. That is, the attractiveness is constant anywhere within the search space. This behavior is a special case of particle swarm optimization (PSO). If $\gamma \rightarrow \infty$, the second term falls out from the Eq.~(\ref{eq:move}), and the firefly movement becomes a random walk, which is essentially a parallel version of simulated annealing. In fact, each implementation of FA can be between these two asymptotic behaviors.

\section{Studies on Firefly Algorithms: Classifications and Analysis }

Several variants of firefly algorithms exist in the literature. A certain classification scheme is necessary in order to classify them. The easiest way to achieve this purpose is to distinguish the firefly algorithms according to the settings of their algorithm parameters (also strategy parameters)~\cite{Eiben:2003}. The settings of these parameters is crucial for good performance and, therefore, carefully selected by developers. In general, there are two ways to seek algorithm parameters properly. On the one hand, by parameter tuning~\cite{Eiben201119} the good values of parameters are found before the algorithm's run, and are fixed during iterations. On the other hand, by parameter control~\cite{Eiben:2003} the values of the parameters are modified during the runs.

Furthermore, the behavior of FA does not depend only on the proper values of parameters, but also on what components or features are incorporated into it. Therefore, the classification scheme should be able to classify firefly algorithms according to these aspects being taken into account. In this review, we were interested in the following aspects:

\begin{itemize}
\item What is modified?
\item How are the modifications made?
\item What is the scope of the modifications?
\end{itemize}

From the first aspect, firefly algorithms can be classified according to the components or features of which they consists. These are:

\begin{itemize}
\item representation of fireflies (binary, real-valued),
\item population scheme (swarm, multi-swarm),
\item evaluation of the fitness function,
\item determination of the best solution (non-elitism, elitism),
\item moving fireflies (uniform, Gaussian, L\'{e}vy flights, chaos distribution).
\end{itemize}

 With regard to the second aspect, the categories of parameter control in firefly algorithms can be divided into: deterministic, adaptive, and self-adaptive. Finally, according to the last aspect, modifications in firefly algorithms may affect: an element of the firefly, the entire firefly, or the whole population.

In the early stages, FA acts as a global problem solver. That is, for several continuous optimization problems, the algorithm found the desired solutions. Difficulty arose, when the appropriate solutions could not be found for some other optimization problems. This is in accordance with the No-Free-Lunch theorem~\cite{Wolpert97nofree}. To circumvent this theorem, hybridization has been applied to optimization algorithms for solving a given set of problems. In line with this, firefly algorithms have been hybridized with other optimization algorithms, machine learning techniques, heuristics, etc. Hybridization can take place in almost every component of the firefly algorithm, for example, initialization procedure, evaluation function, moving function, etc.

\begin{figure*}[htb]  
\begin{center}
\includegraphics [scale=0.7]{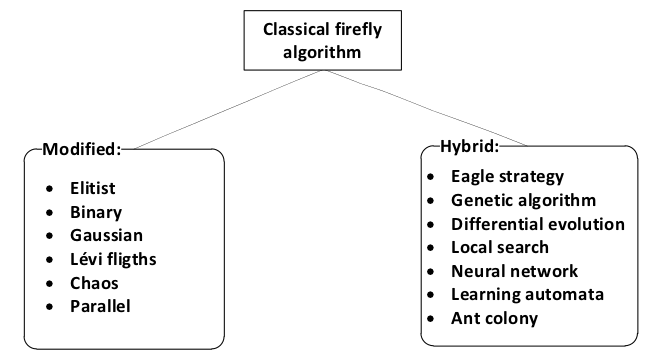}  %
\caption{Taxonomy of firefly algorithms.}
\label{pic:appl}
\end{center}
\vspace{-5mm}
\end{figure*}

In this paper, the firefly algorithms are analyzed according to Figure~\ref{pic:appl}, where the classical firefly algorithms are divided into modified and hybrid. Note that the classical firefly algorithms have been used mainly on continuous optimization problems. In order to provide the optimal results by solving the various classes of problems, they have been subject to several modifications and hybridizations. The main directions of these modifications have gone into the development of: elitist and binary firefly algorithms, Gaussian, L\'{e}vy flights and chaos based firefly algorithms, and the parallelized firefly algorithms. On the other hand, the following hybridizations have been applied to the classical firefly algorithm: Eagle strategy, genetic algorithm, differential evolution, local search, neural network, learning automata and ant colony.

In the rest of this paper, we first present a review of papers describing the classical firefly algorithm. Then, we review the studies that address the modified and hybridized versions of the firefly algorithm. Finally, an overview of papers are carried out that deal with optimization and engineering applications.

\subsection{Classical firefly algorithms}

Firefly algorithm inspired by the variations of light intensity were developed by Yang~\cite{yang2008afirefly} in 2008. In the publication that introduced the classical FA, other nature inspired meta-heuristics were also described. There, it was formulated and the implementation in Matlab was discussed in detail. In order to demonstrate its performance, the four peeks 2-dimensional function was used. The results of those experiments exposed the multi-modal characteristics of this algorithm. That is, the classical FA is able to discover more optimal solutions in the search space simultaneously.

In his paper~\cite{yang2009afirefly}, Yang established that the new FA was especially suitable for multi-modal optimization applications~\cite{Das:2011a}. This claim was derived from the fact that fireflies can automatically subdivide themselves into a few subgroups because neighboring attraction is stronger than long-distance attraction.  Experiments proving this claim were performed for finding the global optima of various multi-modal test functions taken from literature, and compared with genetic algorithm and particle swarm optimization. FA was superior to both the mentioned algorithms in terms of efficiency and success rate. Therefore, the same author speculated that FA is potentially more powerful in solving other NP-hard problems as well.

The same author in~\cite{yang2010firefly} experimented with the classical FA solving the non-linear pressure vessel design optimization problem. This problem belongs to the class of continuous optimization problems. At first, he validated the algorithm using certain standard test functions. The results of FA for pressure vessel design optimization implied that this algorithm is potentially more powerful than other existing algorithms such as particle swarm optimization.

Yang's paper~\cite{yang2011review} provides an overview of the nature inspired meta-heuristic algorithms, including ant colony optimization, cuckoo search, differential evolution, firefly algorithm, harmony search, genetic algorithm, simulated annealing, and particle swarm optimization. On the base of the common characteristics of these, Yang proposed a new, generic, meta-heuristic algorithm for optimization called the Generalized Evolutionary Walk Algorithm (GEWA) with the following three major components: 1) global exploration by randomization, 2) intensive local search by random walk, and 3) the selection of the best using some elitism. Interestingly, this algorithm tries to explicitly balance exploration and exploitation via a new randomization control parameter. However, the value of this parameter is problem-dependent.

Parpinelli, in his survey `New inspirations in swarm intelligence'~\cite{parpinelli2011new}, explained the foundations that inspire developers of the new nature-inspired swarm intelligence algorithms. These foundations are: bacterial foraging (BFO)~\cite{Passino:2002}, fireflies' bioluminescense~\cite{yang2008afirefly}, slime moulds life-cycle~\cite{Monismith:2008}, cockroach infestation~\cite{Havens:2009}, mosquitoes host-seeking~\cite{Feng:2009}, bats echolocation~\cite{Yang:2010b}, and various bees algorithms (BAs), i.e., inspired by bees foraging~\cite{Karaboga:2005}, and bees mating~\cite{Haddad:2004}. The more important applications and the main features of such meta-heuristics were also reported.

Zang et al.~\cite{zang2010review} systematically reviewed and analyzed the following nature-inspired algorithms: ant colony algorithm, bees algorithm, genetic algorithm, and the firefly algorithm. Although the paper mainly focused on the original principles behind these algorithms, their applications were also discussed.

The complete list of classical firefly algorithms is summarized in Table~\ref{tab:classical}.

\begin{center}
\begin{table}[htb]
\small
\begin{tabular}{  p{10cm} | l  }
\hline
Topic & References \\ \hline
Introducing the firefly algorithm & \cite{yang2008afirefly} \\
Multi-modal test functions & \cite{yang2009afirefly} \\
Continuous and combinatorial optimization & \cite{yang2010firefly} \\
Review of nature inspired meta-heuristics & \cite{yang2011review} \cite{parpinelli2011new} \cite{zang2010review} \\
\hline
\end{tabular}
\caption{Classical firefly algorithms}
\normalsize
\label{tab:classical}
\end{table}
\vspace{-5mm}
\end{center}

\subsection{Modified firefly algorithms}

The firefly algorithms depend primarily on the variation of light intensity and the formulation of attractiveness. Both factors allow significant scope for  algorithm improvements (Table~\ref{tab:modified}). For example, Luleseged et al.~\cite{leleseged:hindawi} modified the random movement of the brightest firefly that in some generations, when the current best position does not improve, may decreases its brightness. The proposed modification tries to improve the brightest firefly position generating the $m$-uniform random vectors and moves it in the direction of the best performance. If such a direction does not exist, the brightest firefly stays in its current position. In this case, the brightest firefly is also an elitist solution, because it is never replaced by a best-found solution in the current generation that has lower fitness. Experiments that were done optimizing seven benchmark functions showed that this modified FA outperform its classical predecessor.

Interestingly, a lot of binary firefly algorithms have emerged for solving different classes of problems, e.g.,~\cite{palit2011cryptanalytic} \cite{falcon2011fault} \cite{chandrasekaran2012network}. Palit in~\cite{palit2011cryptanalytic} proposed a binary FA for cryptanalysis in order to determine the  plain text from the cipher text, using the Merkle-Hellman knapsack cipher algorithm~\cite{Behrouz:2007}. Changes to almost all components of the binary FA needs to be performed because of the new representation of fireflies. The results of this algorithm were compared with the results of the genetic algorithm. This comparison showed that the proposed binary FA was much more efficient than the genetic algorithm when used for the same purpose. Next an implementation of the binary FA was developed by Falcon et al.~\cite{falcon2011fault} that uses a binary encoding of the candidate solutions, an adaptive light absorption coefficient for accelerating the search, and problem-specific knowledge to handle infeasible solutions. The empirical analysis was performed by solving the system-level fault diagnostic that is a combinatorial optimization problem. The results of the proposed algorithm when compared with an artificial immune system approach (AIS)~\cite{YangH:2008} and particle swarm optimization (PSO)~\cite{Falcon:2012a}, showed that it outperforms existing techniques in terms of convergence speed and memory requirements. In the paper of Chandrasekaran et al.~\cite{chandrasekaran2012network}, the authors proposed a binary coded FA for solving network and reliability constrained unit commitment (UC) problem~\cite{Carrion:2006}, without satisfying the network security constraints. The effectiveness of the proposed algorithm was demonstrated on 10 units of the IEEE-RTS system. The results of the proposed algorithm were promising when compared with the other techniques reported.

In order to stabilize fireflies' movements, Farahani in~\cite{farahani1gaussian} formulated a new FA that increases convergence speed using Gaussian distribution to move all fireflies to global best in each iteration. Despite the fixed randomization parameter $\alpha$ (also step size), this parameter was modified adaptively in the proposed algorithm. This algorithm was tested on five standard functions. The experimental results showed better performance and more accuracy than the classical firefly algorithm.

Yang in his paper~\cite{yang2011metaheuristic} intended to provide an overview of convergence and efficiency studies of meta-heuristics, and tried to provide a framework for analyzing meta-heuristics in terms of convergence and efficiency. Three well-known heuristics were taken into account: simulated annealing~\cite{Bertsimas:1993}, particle swarm optimization~\cite{Kennedy:1995}, and the firefly algorithm. The impact of randomization methods Gaussian random walk and L\'{e}vy flight on the results of meta-heuristics was also analyzed in this paper. The conclusion was that the most important issue for the newly developed meta-heuristics was to provide a balanced trade-off between local exploitation and global exploration, in order to work better.

Yang~\cite{yang2010afirefly} formulated a new meta-heuristic FA using the L\'{e}vy flights move strategy. Numerical studies and results suggest that the proposed L\'{e}vy-flight FA is superior to particle swarm optimization and genetic algorithms in regard to efficiency and success rate. The paper~\cite{yang2012efficiency} was dedicated to analyzing the convergence and efficiency associated with meta-heuristics like swarm intelligence, cuckoo search, firefly algorithm, random walks, and L\'{e}vy flights. Although the author Yang tried to discover some mathematical foundations for meta-heuristic behavior, he concluded that despite the fact that the newly developed nature-inspired meta-heuristics worked well on average, mathematical understanding of these partly remains a mystery.

Coelho et al. in their paper~\cite{dos2011chaotic} proposed a combination of FA with chaotic maps~\cite{Strogatz:2000} in order to improve the convergence of the classical firefly algorithm. Use of the chaos sequences was shown to be especially effective by easier escape from the local optima. The proposed firefly algorithms used these chaotic maps by tuning the randomized parameter $\alpha$ and light absorption coefficient $\gamma$ in Eq.~(\ref{eq:move}). A benchmark of reliability-redundancy optimization has been considered in order to illustrate the power of the proposed FA using chaotic maps. The simulation results of the proposed FA were compared with other optimization techniques presented in literature and it was revealed that the proposed algorithm outperformed the previously best-known solutions available. On the other hand, Gandomi et al. in~\cite{gandomi2012afirefly} introduced chaos into FA in order to increase its global search mobility for robust global optimization. Using chaotic maps, they tuned attractiveness $\beta_{0}$ and light absorption coefficient $\gamma$ in Eq.~(\ref{eq:move}). The authors analyzed the influence of using 12 different chaotic maps on the optimization of benchmark functions. The results showed that some chaotic FA can clearly outperform the classical FA.

Interestingly, Subotic et al.~\cite{suboticparallelization} developed the parallelized FA for unconstrained optimization problems tested on standard benchmark functions. Both the speed and quality of the results were placed by the authors and as a result, the parallelized FA obtained much better results over much less execution time. Unfortunately, this conclusion was valid only when more than one population was taken into account. Husselmann et al. in~\cite{husselmann2012parallel} proposed a modified FA on a parallel graphical processing unit (GPU) where the standard benchmark functions were taken for comparison with the classic firefly algorithm. They revealed that the results of this parallel algorithm were more accurate and faster than by the original firefly algorithm, but this was only valid for multi-modal functions. As matter of fact, the classical FA is well suited to optimizing unimodal functions as very few fireflies are required and, thus, calculation times are dramatically lower.

\begin{center}
\begin{table}[htb]
\small
\begin{tabular}{  p{9cm} | l  }
\hline
Topic & References \\ \hline
Elitist firefly algorithm & \cite{leleseged:hindawi} \\
Binary represented firefly algorithm & \cite{palit2011cryptanalytic} \cite{falcon2011fault} \cite{chandrasekaran2012network} \cite{farahani2012some} \\
Gaussian randomized firefly algorithm & \cite{farahani1gaussian} \cite{yang2011metaheuristic} \\
L\'{e}vy flights randomized firefly algorithm & \cite{yang2011metaheuristic} \cite{yang2010afirefly} \cite{yang2012efficiency} \\
Chaos randomized firefly algorithm & \cite{dos2011chaotic} \cite{gandomi2012afirefly} \\
Parallel firefly algorithm & \cite{suboticparallelization} \cite{husselmann2012parallel} \\
\hline
\end{tabular}
\caption{Modified firefly algorithms}
\label{tab:modified}
\end{table}
\normalsize
\vspace{-5mm}
\end{center}

\subsection{Hybrid firefly algorithms}

According to the No-Free-Lunch theorem, any two general problem solvers are equivalent when their average performance is compared across all possible problems. That is, they can obtain average results on all classes of problems. Specific heuristics are intended to solve a given set of problems, and normally improve the results of the problem solvers, i.e., heuristics exploit a specific-knowledge of the given problem domains. In fact, these heuristics can also be incorporated into a FA that is a kind of general problem solver. Such hybridized firefly algorithm, in general, improves the results when solving the given problem. In contrast, FA can be also used as a heuristic for hybridizing with other general problem solvers because of its characteristics, i.e., multi-modality and faster convergence.

As the first hybridization of the firefly algorithm, Yang in~\cite{yang2010eagle} formulated a new meta-heuristic search method, called Eagle Strategy (ES), which combines the L\'{e}vy flight search with the firefly algorithm. The Eagle strategy was inspired by the foraging behavior of eagles. These eagles fly freely over their territory in a random manner similar to L\'{e}vy flights~\cite{Brown:2007}. When the prey is seen, the eagle tries to catch it as efficiently as possible. From the algorithmic point of view, the Eagle Strategy consists of two components: random search by L\'{e}vy flight, and intensive local search. Interestingly, FA was applied for the local search. This hybrid meta-heuristic algorithm was employed to the Ackley function with Gaussian noise. The results of simulation showed that the Eagle strategy could significantly outperform the particle swarm optimization algorithm in terms of both efficiency and success rate.

The paper of Luthra et al.\cite{luthra2011hybrid} discussed the hybridization of FA for cryptanalysis of the mono-alphabetic substitution cipher with the operators of mutation and crossover commonly used in Genetic Algorithms. Dominant gene crossover was used as the crossover operator, whilst the permutation mutation was taken into account for mutation. From the experiments, it was observed that the algorithm worked better for large input cipher text lengths. For smaller input cipher lengths, a larger total number of generations would need to be used.

In~\cite{abdullah2012new} Abdullah et. al proposed a Hybrid Evolutionary Firefly Algorithm (HEFA) that combined the classical Firefly Algorithm with the evolutionary operations of the Differential Evolution (DE) method in order to improve searching accuracy and information sharing amongst the fireflies. This algorithm divided the population of fireflies in two sub-populations according to fitness. In the first, the classical firefly operators were applied, whilst in the other, the evolutionary operators were adopted from the Differential Evolution~\cite{Storn:1997} \cite{Brest:2006} \cite{Das:2011}. The proposed method was used to estimate the parameters in a biological model. The experimental results showed that the accuracy and speed performance of HEFA had significantly outperformed the results produced by the genetic algorithms, particle swarm optimization, evolutionary programming, and the classical firefly algorithm.

In the paper of Fister Jr. et al.~\cite{fister2012memetic}, the classical FA was hybridized using local search heuristic and applied to graph 3-coloring that is a well-known combinatorial optimization problem~\cite{fister2012coap}. The results of the proposed memetic FA (MFA) were compared with the results of the Hybrid Evolutionary Algorithm (HEA)~\cite{Hao:1999}, Tabucol~\cite{Hertz:1987}, and the evolutionary algorithm with SAW method (EA-SAW)~\cite{Eiben:1998} by coloring a suite of medium-scaled random graphs (graphs with 500 vertices) generated using the Culberson random graph generator. The results of FA were very promising and showed the potential that FA could successfully be applied to the other combinatorial optimization problems as well.

Hassanzadeh et al.~\cite{hassanzadeh2012speech} used FA for training the parameters of the Structure Equivalent Fuzzy Neural Network (SEFNN) in order to recognize the speech. FA improved the ability of generalizing fuzzy neural networks. The results showed that this hybridized algorithm for speech recognition had a higher recognition rate than the classical fuzzy neural network trained by the particle swarm optimization method. On the other hand, Nandy et al.~\cite{nandy2012analysis} applied a firefly meta-heuristic with back-propagation method to train a feed-forward neural network. Here, the firefly algorithm was incorporated into a back-propagation algorithm in order to achieve a faster and improved convergence rate when training feed-forward neural network. The proposed hybrid algorithm was tested over some standard data sets. It was revealed that the proposed algorithm converged to local optima within a few iterations. These results were compared with the results of the genetic algorithm optimized the same problems. It was observed that the proposed algorithm consumed less time to converge and improved the convergence rate with minimum feed-forward neural network design.

In the paper of Hassanzadeh et al.~\cite{hassanzadehnew}, cellular learning automata were hybridized with the firefly algorithm. In this meta-heuristic, the cellular learning automata were responsible for making diverse solutions in the firefly population, whilst FA improved these solutions in the sense of local search. The performance of the proposed algorithm was evaluated on five well-known benchmark functions. The experimental results showed that it was able to find the global optima and improve the exploration rate of the standard firefly algorithm.

Farahani~\cite{farahani2012some} proposed three classes of algorithms for improving the performance of the classical firefly algorithm. In the first class, learning automata were used for adapting the absorption and randomization parameters in the firefly algorithm. The second class hybridized the genetic algorithm with FA in order to balance the exploration and exploitation properties of this proposed meta-heuristic by time. The last class used random walk based on a Gaussian distribution in order to move the fireflies over the search space. Experimental results on five benchmark functions showed that the proposed algorithms were highly competitive with the classical firefly and the particle swarm optimization algorithm.

Aruchamy et al.~\cite{aruchamy2011comparative} developed the Flexible Neural Tree (FNT) model for micro-array data to predict cancer using the Ant Colony Optimization (ACO). The parameters encoded in the neural tree were tuned using a firefly algorithm. This proposed model helped to find the optimal solutions at a faster convergence to lower error. In extensive experiments, a comparison was performed between FA and exponential particle swarm optimization (EPSO). The results showed that FA was superior to the EPSO in terms of both efficiency and success.

A list of hybrid firefly algorithms is presented in Table~\ref{tab:hybrid}.

\begin{center}
\begin{table}
\small
\begin{tabular}{  p{10cm} | l  }
\hline
Topic & References \\
\hline
Eagle strategy using L\'{e}vy walk & \cite{yang2010eagle} \\
Genetic Algorithms & \cite{luthra2011hybrid} \cite{farahani2012some} \\
Differential Evolution & \cite{abdullah2012new} \\
Memetic algorithm & \cite{fister2012memetic} \\
Neural network & \cite{hassanzadeh2012speech} \cite{nandy2012analysis} \\
Cellular learning automata & \cite{hassanzadehnew} \cite{farahani2012some} \\
Ant colony & \cite{aruchamy2011comparative} \\
\hline
\end{tabular}
\caption{Hybrid firefly algorithms}
\label{tab:hybrid}
\end{table}
\normalsize
\vspace{-5mm}
\end{center}

\subsection{Why Firefly Algorithms are so Efficient}

As the literature of firefly algorithms is rapidly expanding, a natural question is
`why FA is so efficient?'. There are many reasons for its success. By analyzing
the main characteristics of the standard/classical FA, we can highlight the following
three points:
\begin{itemize}
\item FA can automatically subdivide its population into subgroups, due to the fact that
local attraction is stronger than long-distance attraction.
As a result, FA can deal with highly nonlinear, multi-modal optimization problems naturally and efficiently.

\item FA does not use historical individual best $s_i^*$, and there is no explicit global
best $g^*$ either. This avoids any potential drawbacks of premature convergence as those in PSO.
In addition, FA does not use velocities, and there is no problem as that associated with velocity in PSO.

\item FA has an ability to control its modality and adapt to problem landscape by controlling
its scaling parameter such as $\gamma$.
In fact, FA is a generalization of SA, PSO and DE, as seen clearly in the next paragraph.

\end{itemize}

In addition, the standard firefly algorithm can be considered as a generalization to
particle swarm optimization (PSO), differential evolution (DE), and simulated annealing (SA).
From Eq.~(\ref{eq:move}), we can
see that when $\beta_0$ is zero, the updating formula becomes essentially a version
of parallel simulated annealing, and the annealing schedule is controlled by $\alpha$.
On the other hand, if we set $\gamma=0$ in Eq.~(\ref{eq:move}) and set $\beta_0=1$
(or more generally, $\beta_0 \in$ Unif(0,1)), FA becomes a simplified version
of differential evolution without mutation, and the crossover rate is controlled by $\beta_0$.
Furthermore, if we set $\gamma=0$ and replace $s_j$ by the current global best solution
$g^*$, then Eq.~(\ref{eq:move}) becomes a variant of PSO, or accelerated particle swarm optimization,
to be more specific. Therefore, the standard firefly algorithm includes DE, PSO and SA
as its special cases. As a result, FA can have all the advantages of these three algorithms.
Consequently, it is no surprise that FA can perform very efficiently.

\section{Applications of Firefly Algorithms}

\begin{figure*}[htb]  
\begin{center}
\includegraphics [scale=0.5]{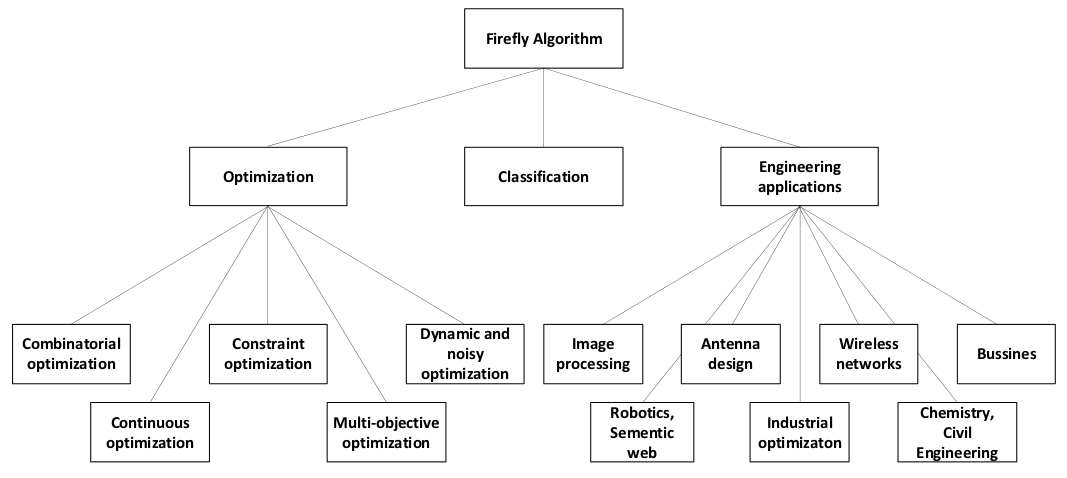}  %
\caption{Taxonomy of firefly applications.}
\label{pic:appl2}
\end{center}
\vspace{-5mm}
\end{figure*}

Nowadays, FA and its variants have been applied for solving many optimization and classification problems, as well as several engineering problems in practice. The taxonomy of the developed firefly algorithm applications is illustrated in Figure~\ref{pic:appl2}. As can be seen from this figure, FA has been applied to the following classes of optimization problems: continuous, combinatorial, constrained, multi-objective, dynamic and noisy optimization. In addition, it has been used for classification problems in: machine learning, data mining, and neural networks. Finally, the firefly algorithms are used in almost all branches of engineering. In this review, we focused on the following engineering areas: image processing, industrial optimization, wireless sensor networks, antenna design, business optimization, robotics, semantic web, chemistry, and civil engineering.

\subsection{Optimization}

\paragraph{Continuous optimization}
Most of classical publications about the firefly algorithm, like~\cite{yang2010firefly} \cite{yang2011metaheuristic} \cite{yang2011review} \cite{yang2010afirefly} and \cite{yang2012efficiency} apply to continuous optimization problems. In most cases, the benchmarks of well-known optimization functions were taken into account. In order to complete the comprehensive picture of this area, the paper of Gandomi et al. in~\cite{gandomi2011mixed} was reviewed, where FA was used for solving mixed continuous/discrete structural optimization problems taken from literature regarding welded beam design, pressure vessel design, helical compression spring design, reinforced concrete beam designs, stepped cantilever beam design, and car side impact design. The optimization results indicated that FA is more efficient than other meta-heuristic algorithms such as particle swarm optimization, genetic algorithms, simulated annealing and differential evolution. Although FA was very efficient, oscillatory behavior was observed as the search process approached the optimum design. The overall behavior of FA could be improved by gradually reducing the randomization parameter as the optimization progressed.

\paragraph{Combinatorial optimization}
FA is also employed for solving combinatorial optimization problems. Durkota in his BSc thesis~\cite{durkota} adjusted FA to solve the class of discrete problems named Quadratic Assignment Problem (QAP), where the solutions are represented as permutations of integers. In this algorithm, the continuous functions like attractiveness, distance and movement, were mapped into newly developed discrete functions. The experimental results were obtained on 11 different QAP problems chosen from the public QAPLIB Library. The author reported fast convergence and success rate on the simplest problems, whilst the algorithm often falls into the local optima by solving the hard problems.

The paper of Sayadi et al.~\cite{sayadi2010discrete} presented a new discrete firefly meta-heuristic for minimizing the makespan for the permutation flow shop scheduling problem that is classified as a NP-hard problem. The results of the proposed algorithm were compared with other existing ant colony optimization technique and indicated that it performs better than the ant colony for some well known benchmark problems.

The first approach to apply the firefly meta-heuristic to the task graph scheduling problem (NP-hard problem) was performed by~\cite{honig2010firefly}. The results on an extensive experimental benchmark with 36,000 task graph scheduling problems showed that the presented algorithm required less computing time than all the other meta-heuristics used in the test, but these were not as promising when the other performance measures were taken into account.

Jati et al.~\cite{jati2011evolutionary} applied FA to the symmetric traveling salesman problem that is a well-known NP-hard problem. In this evolutionary discrete FA (EDFA), a permutation representation was used, where an element of array represents a city and the index represents the order of a tour. \textit{m}-moves were generated for each firefly using inversion mutation. The simulation results indicated that the EDFA performed very well for some TSPLIB instances when compared to the memetic algorithm. Unfortunately, it can often be trapped into local optimum.

Khadwilard et al.~\cite{khadwilard2011application} developed FA for solving the Job shop scheduling problem (JSSP). A computational experiment was conducted using five benchmark datasets of the JSSP instance from the well-known OR-Library~\cite{Beasley:1990}, for finding the lowest makespan. The authors reported that although this FA could found the best known solution in some cases, it was also trapped into the local optima several times.

In~\cite{liu2012new} the authors Liu et al. proposed a path planning adaptive firefly algorithm. Here, the random and absorption parameters were designed to
be adaptive in order to improve the solution quality and convergence speed of the classical firefly algorithm. The simulation tests verified the effectiveness of the improved algorithm and the feasibility of the path-planning method based on the firefly algorithm. In addition, Wang et al.~\cite{wangmodified} developed a new modified FA (MFA) for solving the path planning problem for uninhabited combat air vehicle (UCAV), where a modification is applied for exchanging information between top fireflies during the process of light intensity updating. This modification accelerated the global convergence speed, whilst preserving the strong robustness of the classical firefly algorithm. In order to prove the performance of this algorithm, MFA was compared with classical FA and other population-based optimization algorithms like ant colony optimization (ACO)~\cite{Dorigo:1999}, biogeography-based optimization (BBO)~\cite{Simon:2008}, differential evolution (DE)~\cite{Storn:1997}, evolutionary strategy (ES)~\cite{Back:1996}, genetic algorithm (GA)~\cite{Goldberg:1989}, probability-based incremental learning (PBIL)~\cite{Shumeet:1994}, particle swarm optimization (PSO)~\cite{Kennedy:1999}, and the stud genetic algorithm (SGA)~\cite{Khatib:1998}. The experiment showed that the proposed algorithm was more effective and feasible in UCAV path planning than the other algorithms.

The paper of Kwiecien et al.~\cite{kwiecien2012firefly} experimented with FA for optimization of the queuing system that also belongs to a class of NP-hard problems. The queuing theory provides methods for analyzing complex service systems in computer systems, communications, transportation
networks and manufacturing. This algorithm was tested by solving two problems found in literature. The results of the experiments performed for selected queuing systems were promising when comparing with the results of genetic algorithms.

\paragraph{Constraint Optimization}

The first contribution to the domain of constraint optimization was made by Lukasik et al. in~\cite{lukasik2009firefly} who experimented with the classical FA for constrained continuous optimization. The proposed algorithm was tested by solving a set of 14 benchmark constraint problems and the obtained results were compared with the existing particle swarm optimization algorithm~\cite{Kennedy:1999}. Here, FA was outperformed by particle swarm optimization in their tests.
Later, Gandomi et al.~\cite{gandomi2011mixed} studied a set of nonlinear constraint optimization problems in engineering and showed that
FA can provided better results than other methods such as particle swarm optimization.

Gomes in~\cite{gomes2012firefly} employed FA for optimizing a structural mass optimization on shape and size by taking dynamic constraints into account. This optimization problem had extreme non-linear behavior with regard to the frequency constraints especially for shape optimization, since eigenvalues are very sensitive to shape modifications. In this algorithm, the constraints were treated as penalty functions that affected the fitness function. It was tested by solving three examples of increasing difficulty, which were compared with the results in the literature. From an engineering point of view, FA performed well in all three cases.

\paragraph{Multi-objective optimization}

The contribution of FA to multi-objective optimization~\cite{Zhou:2011} is as follows. Yang in~\cite{yang2012multiobjective} formulated a new multi-objective FA for multi-objective optimization that extended FA for producing Pareto optimal front directly. This algorithm was tested on a subset of multi-objective functions with convex, non-convex, and discontinuous Pareto fronts taken from literature and was then applied for solving design optimization benchmarks in industrial engineering. The results were compared with other algorithms for multi-objective optimization like the vector evaluated genetic algorithm (VEGA)~\cite{Schaffer:1985}, non-dominated sorting genetic algorithm-II (NSGA-II)~\cite{Deb:2002}, multi-objective differential evolution (MODE)~\cite{Babu:2007}, differential evolution for multi-objective optimization (DEMO)~\cite{Robic:2005}, multi-objective bees algorithms (Bees)~\cite{Pham:2007} and strength Pareto evolutionary algorithm (SPEA)~\cite{Zitzler:1999}, and suggested that the proposed algorithm is an efficient multi-objective optimizer.

Abedinia et al. in~\cite{abedinia2012multi} developed a multi-objective FA for an Environmental/Economic Power Dispatch (EED) problem. This problem was formulated as a non-linear constrained multi-objective problem with the conflict objectives of fuel cost, emission, and system loss. The proposed algorithm run on the IEEE 30- and 118-bus test systems and the results were compared with other known multi-objective algorithms from literature. The achieved numerical results of the proposed FA demonstrated the feasibility of this for solving the multi-objective EED problem.

A slightly different problem was solved by Niknam et al. in~\cite{niknam2012new}, where the authors proposed a multi-objective FA in order to achieve a set of non-dominated (Pareto-optimal) solutions. A tuning of randomization and absorption coefficients based on the usage of chaotic maps and self-adaptive probabilistic mutation strategies were used to improve the overall performance of the algorithm. This algorithm solved the problem of dynamic economic emission dispatch (DEED) that incorporated combined heat and power units in so called CHP (Combined Heat and Power) systems. The problem was defined as multi-objective and the performance of the proposed algorithm was successfully validated using numerical simulations.

\paragraph{Dynamic environment}

Many real-world problems are mostly time varying optimization problems. Therefore, these require special mechanisms for detecting changes in the environment and then respond to them. FA can be used also for this dynamic environment. Here, a short review is made of those papers that deal with this class of optimization.

Abshouri et al. in~\cite{abshouri2011new} hybridized FA with learning automata that were responsible for tuning the firefly parameters, i.e., randomized $\alpha$, attractiveness $\beta_{0}$ and absorption coefficient $\gamma$. The main idea in this algorithm was to split the population of fireflies into a set of interacting swarms that were able to respond quicker to the changing environment. This algorithm was tested on a variety of instances of the multi-modal dynamic moving peaks benchmark~\cite{Branke:1999} and the obtained results were compared with other algorithms for dynamic environments like multi-quantum swarm optimization (mQSO)~\cite{Blackwell:2006}, fast multi-swarm optimization (FMSO)~\cite{Li:2008}, and cellular particle swarm optimization~\cite{Hashemi:2009}. They showed that this FA significantly outperformed other mentioned algorithms.

In~\cite{chai2011bees}, Chai et al. adopted two meta-heuristics: bees and firefly algorithms, in order to find optimal solutions to various types of noisy continuous mathematical functions, using two variables. The results were measured with regard to mean and standard deviations of execution time needed to find an optimal solution. According to the obtained results, the authors concluded that FA performed better than the bees algorithm when the noise levels increased.

Farahani et al. in~\cite{farahani2011multiswarm} developed a multi-swarm based FA for dynamic environments that splits the population of particles into sets of interacting swarms. Each swarm interacts locally by an exclusion parameter and globally through an anti-convergence operator. This algorithm was tested solving the moving peaks benchmark~\cite{Branke:1999}. The results were compared according to performance and accuracy with other particle swarm optimization (PSO) and evolutionary algorithms from the literature and showed that the proposed algorithm significantly outperformed other algorithms in experiments.

FA for dynamic environments that was proposed by Nasiri et al. in~\cite{nasiri2012speciation}, had the following characteristics: The best solution found so far was preserved. The randomization parameter $\alpha$ was adapted. The algorithm maintained only one swarm of fireflies.  It was applied to a moving peaks benchmark problem~\cite{Branke:1999}, which is the most famous benchmark for assessment in dynamic environments, and compared with other algorithms for dynamic optimization, like mQSO~\cite{Blackwell:2006}, AmQSO~\cite{Blackwell:2008}, mCPSO~\cite{Blackwell:2006}, SPSO~\cite{du:2008}, rSPSO~\cite{bird:2007} and PSO-CP~\cite{liu:2010}. The obtained results showed the proper accuracy and convergence rate for the proposed algorithm in comparison with the other mentioned algorithms.

Finally, Sulaiman in \cite{sulaiman2012optimal} presented an application of FA for determining the optimal location and size of distributed generation (DG) in distribution power networks. In this paper, an IEEE 69-bus distribution test system was used to show the effectiveness of the firefly algorithm. The comparison with the genetic algorithm was also conducted to see the performance of FA as to whether it was as good as the genetic algorithm in solving the optimal allocation problem.

The detailed list of optimization algorithms can be seen in Table~\ref{tab:optimization}.

\begin{center}
\begin{table}[htb]
\small
\begin{tabular}{  p{6cm} | p{6.5cm}  }
\hline
Topic & References \\
\hline
Continuous optimization & \cite{yang2010firefly} \cite{yang2011metaheuristic} \cite{yang2011review} \cite{yang2010afirefly} \cite{yang2012efficiency} \cite{gandomi2011mixed} \\
Combinatorial optimization & \cite{durkota} \cite{sayadi2010discrete} \cite{honig2010firefly} \cite{jati2011evolutionary} \cite{khadwilard2011application} \cite{liu2012new} \cite{wangmodified} \cite{kwiecien2012firefly} \\
Constrained Optimization & \cite{lukasik2009firefly} \cite{gandomi2011mixed} \cite{gomes2012firefly} \\
Multi-objective optimization & \cite{yang2012multiobjective} \cite{abedinia2012multi} \cite{niknam2012new} \\
Dynamic and noisy environment & \cite{abshouri2011new} \cite{chai2011bees} \cite{nasiri2012speciation} \cite{farahani2011multiswarm} \cite{sulaiman2012optimal} \\
\hline
\end{tabular}
\label{tab:optimization}
\caption{Optimization applications}
\end{table}
\normalsize
\vspace{-5mm}
\end{center}

\subsection{Classification}

Classification algorithm is a procedure for selecting a hypothesis from a set of alternatives that best fits a set of observations or data. Usually, this kind of algorithm appears in machine learning, data mining, and neural networks. Although classifications can be considered as optimization, Holland~\cite{Holland:1992} wrote that a learning (as component part of classification) is viewed as a process of adaptation to a particularly unknown environment, not as an optimization problem. In the rest of this subsection, contributed papers from this area are reviewed.

Banati et al. in~\cite{banati2011fire} hybridized FA with the Rough Set Theory (RST) to find a subset of features and is a valuable preprocessing techniques in machine learning. FA simulated the attraction system of the real fireflies that guides the feature selection procedure. Four different medical datasets from the UCI machine learning data repository~\cite{website:Asuncion} were evaluated in order to obtain the performance of the proposed algorithm. The experimental results showed that the proposed algorithm outperformed the other feature selection method like Genetic algorithm (GenRSAR)~\cite{liu2009nature}, Ant colony Optimization (AntRSAR)~\cite{jensen2003finding}, particle swarm optimization (PSO-RSAR)~\cite{yue2007new} and artificial bee colony (BeeRSAR)~\cite{suguna2010novel}, in terms of time and optimality.

Horng et al. in~\cite{horngfirefly1} proposed FA for training of the radial basis function (RBF) network that is a type of neural network using a radial basis function as its activation function~\cite{Ou:2005}. In these algorithms, each of the fireflies represented a specific RBF network for classification. The performance of the proposed FA was measured according to the percent of correct classification, and the mean square error and complexity on five data sets was taken from the UCI machine repository~\cite{website:Asuncion}. The obtained result for the firefly algorithms were compared with the gradient descent algorithm (GD)~\cite{Karayiannis:1999}, the genetic algorithm (GA), the particle swarm optimization (PSO), and the artificial bee colony (ABC) algorithm.  The experimental results showed that usage of FA obtained satisfactory results over the GD and GA algorithms, but it was not apparently superior to the PSO and ABC algorithms.

In paper~\cite{senthilnath2011clustering}, Senthilnath et al. applied FA for clustering data objects into groups according to the values of their attributes. The performance of FA for clustering was compared with the results of other nature-inspired algorithms like artificial bees colony (ABC)~\cite{Karaboga:2007} and particle swarm intelligence (PSO)~\cite{Kennedy:1999}, and and other nine methods used in the literature on the 13 test data sets from literature~\cite{karaboga:2010} \cite{falco:2007}. The performance measure used in the comparison was the classification error percentage (CEP) that is defined as a ratio of the number of misclassified samples in the test data set and total number of samples in the test data set. The authors concluded from the obtained results that FA was the efficient method for clustering.

\subsection{Engineering applications}

Firefly algorithms have become a crucial technology for solving problems in engineering practice. Nowadays, there are applications for almost every engineering areas. Table~\ref{tab:EA} shows those engineering areas in which FA was applied with reference to the paper in which the algorithm is described, and the summarized number of papers from a given area. These papers were not analyzed explicitly because of this reviews' scope.

The greatest number of papers in Table~\ref{tab:EA} belong to industrial optimization. Image processing is also well covered by firefly algorithms. Antenna optimization consists of four papers. In business optimization, robotics, and civil engineering, there are two papers for each area. Other engineering areas in the table are represented by one paper.

In summary, every day new papers emerge using FA within always new engineering areas. This fact proves that the development of engineering applications using firefly algorithms is very diverse and rapidly expanding.

\begin{center}
\begin{table}[htb]
\small
\begin{tabular}{  p{4.5cm} | p{7cm} | p{1cm}  }
\hline
Engineering area & References & Total \\
\hline
Industrial Optimization & \cite{apostolopoulos2010application} \cite{yang2011afirefly} \cite{jeklene2011optimization} \cite{chatterjee2012design} \cite{kazemzadeh2011optimum} \cite{aungkulanon2011simulated} \cite{rampriya2010unit} \cite{chandrasekaran2011demand} \cite{hu2012fa} \cite{dekhici2012firefly} \cite{roevafirefly} \cite{roeva16optimization} \cite{abediniafuzzy} \cite{dutta2011exploring} & 14 \\
Image processing & \cite{zhang2012novel} \cite{horng2010codebook} \cite{horng2012vector} \cite{hassanzadeh2011non} \cite{hassanzadeh2011image} \cite{mohd2011multilevel} \cite{horng2010multilevel} \cite{horng2011multilevel} & 8 \\
Antenna design & \cite{basu2011fire} \cite{zaman2012nonuniformly} \cite{chatterjee2012minimization} \cite{basu2012thinning} & 4 \\
Business optimization & \cite{yang2011accelerated} \cite{giannakouris2010experimental} & 2 \\
Civil engineering & \cite{talatahari2012optimum} \cite{gholizadeh2012comprative} & 2\\
Robotics & \cite{jakimovski2010firefly} \cite{severin2012comparison} & 2\\
Semantic web & \cite{pop2011hybrid} & 1\\
Chemistry & \cite{fateen2012evaluation} & 1 \\
Meteorology & \cite{dos2012firefly} & 1 \\
Wireless sensor networks & \cite{breza2008lessons} & 1 \\
\hline
\end{tabular}
\caption{Engineering applications}
\label{tab:EA}
\end{table}
\normalsize
\vspace{-5mm}
\end{center}

\section{Discussion and further work}

This section analyses the main characteristics of the firefly algorithm. That is, the firefly algorithms are suitable for multi-modal optimization, they have fast convergence, obtain good results on function optimization, and are appropriate for combinatorial optimization. The reason for the fast convergence is loosing the diversity of population. In order to keep using this algorithm for large-scale optimization problems, a balancing of the exploration and exploitation, as explained by Yang in~\cite{yang2011review}, should be established.

The results of firefly algorithms depend on the best-found solution within a swarm. Therefore, improving the best solution can improve the search power of the swarm. Experiments using binary firefly algorithms show that this algorithm could successfully be applied to those problems where binary representation is necessary. In some cases, the convergence speed can be improved, when Gaussian or L\'{e}vy flight searches are taken in the move function. Using chaotic maps is appropriate for tuning parameters. Parallel firefly algorithms with one swarm improve the results for solving multi-modal functions, whilst the results are deteriorated on the unimodal functions. In general, parallelism is worthwhile, when the multi-swarm population scheme is employed.

Analyses of papers hybridizing the firefly algorithms show that these are more used as a local search for hybridizing other algorithms, like Eagle Strategy, neural networks, learning automata and genetic algorithms. Amongst other approaches, it is worth mentioning the hybridization of firefly algorithms with neural network that acts as meta-heuristic for setting parameters of firefly algorithm, and the hybridization of FA with local search heuristics. In the latter case, FA acts as a global problem solver. Interestingly, there also exists hybridizations that supports the co-evolution of genetic algorithms and firefly algorithms, and operators borrowed from genetic algorithms. However, there are a lot of possibilities for future developments.

Solving the combinatorial optimization problems illustrates the oscillatory behavior of the firefly algorithm. That is, FA can find the solution faster in some cases, whilst in other cases the solution cannot be found. Here, the main problem to be solved is how to balance the exploration and exploitation during firefly search. This question also opens up new directions in the further development of this algorithm.

Using firefly algorithms for combinatorial optimization problems demands a mapping of the discrete variables (usually permutation of integer) in the problem space to continuous variables in the search space, where the firefly operators (like moving fireflies) act. Normally, firefly algorithms for solving the constrained optimization problems consider the constraints as a penalty function that punishes the infeasible solutions. This violation affects the fitness function, this means, the more the solution violates the feasibility condition the higher the value of the penalty function.

Dynamic environment is usually solved by FA using the multi-swarm population scheme, which can respond quicker to the changing environment. Additionally, the adaptation to changing environment is faster, when the control parameters of FA are taken into account. Normally, the learning automata was employed to hybridize FA for dynamic environments. The adaptation of control parameters is also useful for solving the multi-objective problems, whilst for classification problems, despite the adaptation of control parameters, the hybridization of FA is applicable with other classification heuristics like neural networks.

Despite the huge success of FA in practice, mathematical analysis of the algorithm has a very limited literature. In real applications, firefly algorithms often converge quickly, however, there is no theoretical analysis how quickly it can indeed converge. It will be very useful to use theories of dynamical
systems, and/or Markov chains as well as complexity to analyze the convergence and stability of all major variants of firefly algorithm. In fact, there are a few important open problems concerning all meta-heuristic algorithms \cite{yang2011metaheuristic}.

Another important area is parameter tuning, which is important for all meta-heuristic algorithms. In firefly algorithms, more studies are needed to identify optimal setting for algorithm-dependent parameters so that they can solve a wider range of problems with minimal adjustments of parameter values. This itself is a tough optimization problem. It is possible to even design automatic schemes to tune parameters in an adaptive and intelligent manner, but how to achieve such goals is still an open question.

Applications of firefly algorithms are very diverse, as we have seen from this review. It would be more fruitful to apply to new areas such as bioinformatics, data mining, telecommunications, and large-scale real-world applications~\cite{Zamuda:2012}. There is no doubt that more applications of firefly algorithms will emerge in the near future.

\section{Conclusions}

FA algorithm has widely expanded its application domains since its establishment in 2008. Nowadays, there is practically no domain where FA had not been applied. Moreover, the development areas of this algorithm are very dynamic, because new applications appear almost every day.

This paper performs a comprehensive review of the firefly algorithm. It has shown that this algorithm:

\begin{itemize}
\item possesses multi-modal characteristics,
\item can handle multi-modal problems efficiently,
\item has a fast convergence rate,
\item can be used as a general, global problem solver as well as a local search heuristic,
\item is applicable to every problem domain.
\end{itemize}

FA solves problems from the following domains: continuous optimization, combinatorial optimization, constraint optimization, multi-objective optimization, as well as dynamic and noisy environments, and even classification. The important domain of its applicability is represented by engineering application.

The objective of this review was twofold. Firstly, it has summarized the status of applications within various application areas. This review has proved that FA can be practically applied within every problem domain. Secondly, this review has shown that FA is simple, flexible and versatile, which is very efficient
in solving a wide range of diverse real-world problems. At the same, we have provided some open questions and challenges that can inspire further research in these areas in the near future, like ensemble of parameters in~\cite{mallipeddi2011differential}.




\small
\bibliographystyle{elsarticle-num}
\bibliography{bibtex}
\normalsize






\end{document}